\documentclass[preprint,11pt]{elsarticle}

\usepackage{latexsym}
\usepackage{amsmath}
\usepackage{amsfonts}
\usepackage{dsfont}
\usepackage{verbatim}
\usepackage{hyperref}
\usepackage{subcaption}
\usepackage{amsmath,amssymb}
\usepackage{graphicx}
\usepackage{color}
\usepackage{enumerate}
\usepackage{booktabs}
\usepackage{multirow}
\usepackage{url}
\usepackage[shortlabels]{enumitem}
\usepackage{notoccite}
\usepackage[ruled,vlined]{algorithm2e}
\usepackage{xcolor,colortbl}
\usepackage{placeins}
\usepackage{sidecap}

\def\be{\begin{equation}}
\def\ee{\end{equation}}
\def\bea{\begin{eqnarray}}
\def\eea{\end{eqnarray}}

\begin{document}

\begin{frontmatter}

\title{Deep Evidential Learning for
Radiotherapy Dose Prediction}


\author[inst1,inst3]{Hai Siong Tan\footnote{Corresponding Author. Email: haisiong.tan@gryphonai.com.sg}}
\author[inst2]{Kuancheng Wang}
\author[inst3]{Rafe McBeth}

\affiliation[inst1]{Gryphon Center for Artificial Intelligence and 
Theoretical Sciences,
            country={Singapore}}
            
\affiliation[inst2]{Georgia Institute of Technology, Atlanta, GA,
            country={USA}}

\affiliation[inst3]{University of Pennsylvania, Perelman School of Medicine, Department of Radiation Oncology, Philadelphia,
            country={USA}}

\begin{abstract}
\emph{Background}: As we navigate towards integrating deep learning methods in the real clinic, a safety concern lies in whether and how the model can express its own uncertainty when making predictions.
In this work, we present a novel application of an uncertainty-quantification framework called Deep Evidential Learning in the domain of radiotherapy dose prediction.
\\[2ex]
\emph{Method}: 
Using medical images of the Open Knowledge-Based Planning Challenge dataset, we found that this model can be effectively harnessed to yield uncertainty estimates 
that inherited correlations with prediction errors upon completion of network training. This was achieved only after reformulating the original loss function for a stable implementation.
\\[2ex]
\emph{Results}:
We found that (i)epistemic uncertainty was highly correlated with prediction errors, with various association indices comparable or stronger than those for Monte-Carlo Dropout and Deep Ensemble methods, (ii)the median error varied with uncertainty threshold much more linearly for epistemic uncertainty in Deep Evidential Learning relative to these other two conventional frameworks, indicative of a more uniformly calibrated sensitivity to model errors, 
(iii)relative to epistemic uncertainty, aleatoric uncertainty demonstrated a more significant shift in its distribution in response to Gaussian noise added to CT intensity, compatible with its interpretation as reflecting data noise. 
\\[2ex]
\emph{Conclusion}:
Collectively, our results suggest that Deep Evidential Learning is a promising approach that can endow deep-learning models in radiotherapy dose prediction with statistical robustness. We have also demonstrated how this framework leads to uncertainty heatmaps that correlate strongly with model errors, and how it can be used to equip the predicted Dose-Volume-Histograms with confidence intervals.  

\end{abstract}

\end{frontmatter}



\section{Introduction}

Radiotherapy treatment planning is a highly intricate process requiring
collaborative interactions among medical physicists, dosimetrists and radiation oncologists in formulating  the optimal treatment plan (see e.g. \cite{woody, aapm}) for each individual patient. It is a procedure that is characterized by a potentially large degree of user-variability arising from different inter-institutional guidelines, the fundamental origin of which lies in the subjective nature of tradeoffs like those between tumor control and sparing of normal tissues, etc. \cite{woody,Nelms,Craft} Automation by machine learning methods can help to homogenize these variations among the medical professionals, while attaining improvements in the consistency of overall plan quality (see e.g. \cite{Liu}). The adoption of artificial intelligence (A.I.) related techniques led to what is known as knowledge-based planning (KBP) \cite{Wu}, which leverages knowledge inherent in past clinical treatment plans to generate new ones with minimal intervention from human experts. 

A complete KBP method can normally be regarded as a two-stage pipeline \cite{Babier2} : (i)prediction of the dose distribution that should be delivered to patient (ii)conversion of the prediction into a deliverable treatment plan via optimization. Recent dose prediction models cover quite a range of 
sites and modalities \cite{Liu, Rafe}, including
prostate intensity-modulated-radiation-therapy (IMRT) \cite{N_2017, Kearney}, 
prostate volumetric modulated arc therapy (VMAT)  \cite{Shiraishi}, lung
IMRT \cite{Barragan} and head and neck VMAT 
\cite{N_2019b}. These models are designed to predict volumetric dose distributions from which dose-volume histogram (DVH) and other dose constraint-related statistics can be deduced. 
Within the literature, it has been noted \cite{Wu}
that the vast majority of published works was performed using large private datasets which made comparison of model quality challenging. Towards alleviating this issue, the Open Knowledge-Based Planning (OpenKBP) Grand Challenge was organized \cite{Babier2,Babier} to enable an international effort for the comparison of dose prediction models on a single open dataset involving 340 patients of head and neck cancer treated by IMRT. To date, there has been many excellent dose prediction models
trained on the OpenKBP dataset as reviewed
in \cite{Liu}. Notable examples include
\emph{DeepDoseNet} of \cite{L95} which combined features of \emph{ResNet} and \emph{DenseNet}, 
the \emph{MtAA-NET} of \cite{L39} which is based on a generative adversarial network, and \emph{TrDosePred} of \cite{L15} and \emph{Swin UNETR++} of \cite{harry} which are Transformer-based frameworks.\footnote{For a more complete list, see Table 5 of \cite{Liu}.}. 
Like the above-mentioned projects, our work here leverages upon the OpenKBP dataset and attempts to furnish dose prediction. However, unlike the majority of these papers, the major focus of our work here lies in developing a dose prediction model that also encapsulates an \emph{uncertainty quantification} framework. 

As we navigate towards integrating deep learning methods in the real clinic, a safety-related concern lies in whether and how the model can express its own uncertainty when making predictions
based on real-life data distributions beyond those that they were trained on. Uncertainty quantification has been a major theme in the broader context of machine learning \cite{abdar}, and has recently gathered increasing attention in the domain of medical image analysis \cite{lambert}. However, to our knowledge, there is a scarcity of works devoted specifically to uncertainty quantification in dose prediction models. As already noted in papers dedicated to medical image analysis \cite{lambert,ghesu,chexpert,zou,Jones}, uncertainty estimates equip deep learning models with statistical robustness and a measure of reliability that can be useful in discovering regions of the model's ignorance, thereby alerting the human expert to potentially erroneous predictions.
In the domain of dose prediction, 
uncertainty estimates can be used
to characterize the reliability of a model. In the KBP pipeline, dose prediction models are themselves inputs to some dose mimicking model \cite{Babier2} which generates deliverable treatments via optimization of a set of objective functions \cite{Benson,Chan}. Apart from using reliability as a selection criterion for various dose prediction models, it was pointed out in \cite{Babier2,Zhang} that probabilistic dose distributions can serve as robust inputs to the objective functions. At the time of writing, we are only aware of \cite{Rafe} and \cite{Zhang0} being the only publications that examined the role of uncertainty quantification frameworks for dose prediction. The work in \cite{Rafe} studied how Monte-Carlo Dropout (MC Dropout)
and a Deep Ensemble-based bagging method can furnish useful uncertainty estimates for a U-Net-based dose prediction model, whereas \cite{Zhang0} proposed a Gaussian mixture model in which the standard deviation of each Gaussian mixture is part of the network's outputs and can be used to reflect the tradeoff implicit in data from two different treatment protocols.

In this work, we examined the applicability of a relatively new uncertainty quantification framework known as \emph{Deep Evidential Learning} \cite{amini,sensoy} for dose prediction. Its fundamental principle lies in asserting a higher-order Bayesian prior over the probabilistic neural network output, embedding it within the loss function and subsequently generating uncertainty estimates together with the completion of model training. There are two major variants of it as proposed in the seminal works of 
Sensoy et al. in \cite{sensoy} [`\emph{Deep
Evidential Classification}']
and Amini et al. in \cite{amini} [`\emph{Deep
Evidential Regression}'].
We apply the formalism of \cite{amini} to construct a dose prediction model with 3D U-Net \cite{olaf} as the backbone model architecture. In the process, we found that this approach required additional crucial refinements for an effective and stable implementation. 
These refinements are related to the form of the final model layer which connects inputs and weights to the parameters of the higher-order Bayesian prior distribution.
These parameters are formulated as the model's outputs, and are thus naturally obtained when training is completed. They are then used to furnish estimates of model uncertainty. 
This is in contrast to two major categories of uncertainty quantification frameworks in the current literature\footnote{A general survey can be found in \cite{abdar} whereas comprehensive recent reviews in the field of medical 
imaging can be found in \cite{lambert} and \cite{zouReview}. }:
(i) MC Dropout (ii) Deep Ensemble method.
For MC Dropout \cite{gal}, one typically inserts stochastic dropout variables at various points of the neural networks, and after training, a number of 
feedforward passes are performed to evaluate the mean and variance of the output. The latter is then regarded as the model uncertainty. For Deep Ensemble method \cite{lak,Ganaie}, one typically collects a family of models sharing the main network structure but differing in hyperparameters or simply the initial random weight distributions. From the ensemble set of outputs, the model prediction is taken as the average with uncertainty being the variance. 

A prominent feature of Deep Evidential Learning is that the higher-order distribution parameters can be used to compute two different types of uncertainties: the aleatoric and epistemic uncertainties. The former probes the level of noise in data while the latter characterizes the uncertainty intrinsic to the model (see
e.g. \cite{kendall} for a detailed explanation). 
Distinguishing between these two classes of uncertainties is useful since aleatoric uncertainty is largely indicative of noise level in data whereas epistemic uncertainty points to model insufficiency to generalize beyond the training dataset, be it a problem of data insufficiency itself or an unsuitable order of model complexity for the task, etc. In our work, we demonstrated that these two uncertainty types can indeed be differentiated by a noise-sensitivity test and the degree to which each was associated with prediction errors.

Like in \cite{Rafe,L95,L39,L15}, our study is based on the OpenKBP dataset, and involved constructing a dose prediction model as part of the KBP pipeline. As emphasized in \cite{Babier2}, although
the mean absolute error (MAE) is an indicator of the model's feasibility as a dose prediction system, one should bear in 
mind that dose prediction models enact an intermediate role in the complete KBP pipeline. Ultimately, it is the deliverable treatment plan most compatible with various clinical criteria that we wish to generate. For the OpenKBP Challenge, while prediction models with better MAE dose score generally led to treatment plans with higher criteria satisfaction, the best KBP treatment plan in \cite{Babier2} turned out to be associated with the one that ranked 16$^{\text{th}}$ on the dose score chart with a MAE of $\sim 3.19$ Gy. In comparison, our Deep Evidential model's MAE score was $3.09$ Gy. 
We also found that for a stable and effective implementation of the Deep Evidential model, much fine-tuning of model hyperparameters was required. Thus, we picked a vanilla 3D U-Net as our backbone architecture of which learning curve approximately flattened within a couple of hours' training. This enabled us to perform extensive ablation experiments for completing a reformulation of the original theory in \cite{amini}, so that it can be efficiently adapted for dose prediction with excellent uncertainty-error correlation. The primary focus of our work here is to elucidate the extent to which uncertainty estimates are correlated with model prediction errors, as they can be further harnessed to characterize reliability of a model. 

Our paper is organized as follows. 
In Section \ref{sec:P}, we furnish a brief exposition of the theoretical basis underpinning Deep Evidential Learning \cite{amini,sensoy}, our reformulation of the loss function and some brief comments on related works in \cite{Rafe,Zhang0}. In Section \ref{sec:M}, we outline details of our model structure and implementation. This is followed by a presentation of various results in Section \ref{sec:Results}. We end with a summary of key findings, main limitations and significance of our work in Sections \ref{sec:D} and \ref{sec:C}.


\section{Preliminaries}
\label{sec:P}
\subsection{Model uncertainties from a Bayesian prior
probability distribution}

In the standard supervised learning formalism, typically the model is trained via minimizing a loss function such as the mean squared error so that the neural networks' weights converge to at least a local minimum point, yielding a good accuracy defined in terms of the error term implied by the loss function. This standard approach does not directly give any estimates of model or data noise uncertainties. In a more refined approach, one can attempt to provide an estimate of uncertainty by the following assertion: 
\begin{itemize}
\item The target outputs can be modeled as being drawn from a probability distribution. For example, in our context of dose prediction, we let the predicted dose $y_k$ in voxel $V_k$ be described by a Gaussian probability distribution function (PDF) with mean and variance parameters $(\mu_k, \sigma_k )$. 
\item 
We then attempt to train the model to infer 
the set of $(\mu_k, \sigma_k )$ (for every voxel $k$) 
by the principle of maximum likelihood estimation. Typically, this is performed by minimizing the
negative log-likelihood function $\mathcal{L}$ defined as
\be
\label{Gaussian}
f(y_k | \mu_k, \sigma_k ) = \frac{1}{\sqrt{2\pi \sigma^2_k}} e^{-\frac{(y_k - \mu_k )^2}{2\sigma^2_k}},
\,\,\,\,\,\,
\mathcal{L}(\vec{\mu}, \vec{\sigma} )
=-\log \left( \Pi_k f(y_k | \mu_k, \sigma_k ) \right).
\ee
For an ensemble of models being trained upon the same dataset to yield estimates of the
parameters $(\mu_k, \sigma_k)$, one would obtain
a spectrum of differing values of $(\mu_k, \sigma_k)$,
the distribution being stochastic in nature due to distinct initial weight distributions, and their sensitivity to the random noise present in the training dataset. 
\end{itemize}

The Deep Evidential Learning framework of \cite{sensoy} and \cite{amini} posits that
there is another PDF which describes the distribution of the Gaussian mean and variance parameters. This can be regarded as a `higher-order' PDF that in turn describes
the distribution of $(\mu_k, \sigma_k)$
of the `first-order' PDF $f(y|\mu, \sigma)$
in eqn. \eqref{Gaussian}. 
In Bayesian theory, there is a natural mathematical entity that enacts this role -- the prior probability distribution
that treats $(\mu_k, \sigma_k)$ as random rather than deterministic variables. 
In \cite{amini}, various model training and simulations were performed by taking the prior distribution to be a product of a Gaussian distribution (for $\mu_k$)
and an inverse-gamma distribution (for $\sigma_k$). 
Each of them is separately a two-parameter distribution.
Suppressing the voxel index $k$, we have
\bea
&&\mu \sim \mathcal{N} (\gamma, \sigma^2/\nu ), \,\,\,
p(\mu| \gamma, \sigma^2/\nu ) = 
\frac{\sqrt{\nu}}{\sqrt{2\pi \sigma^2}} e^{-\frac{\nu(\gamma - \mu )^2}{2\sigma^2}},\cr
\label{mu_sigma}
&&\sigma^2 \sim \Gamma^{-1}(\alpha, \beta ),\,\,\,
p(\sigma | \alpha, \beta) = 
\frac{\beta^\alpha}{\Gamma(\alpha) \sigma^{2(\alpha + 1)}} e^{-\frac{\beta}{\sigma^2}},
\eea
where $\Gamma^{-1}(x)$ is the inverse gamma distribution.
The overall prior distribution is a four-parameter
\emph{normal-inverse-gamma}
distribution defined as the product of the two PDFs above. 
\be
\label{fNIG}
\mathcal{P}(\mu, \sigma | \alpha, \beta, \nu, \gamma ) =
p(\mu| \gamma, \sigma^2/\nu )p(\sigma | \alpha, \beta) =
\frac{\beta^\alpha \sqrt{\nu}}{\Gamma(\alpha ) \sqrt{2\pi \sigma^2}} \left(  
\frac{1}{\sigma^2}
\right)^{\alpha + 1}
\text{exp}\left[ - \frac{2\beta + \nu (\gamma - \mu)^2}{2\sigma^2} 
\right].
\ee
Given this prior distribution, we can take expectation values and other statistical moments with respect to it to compute mean and variances. In the simplest form, Deep Evidential Learning implies attaching a final feedforward layer (e.g. of the convolutional type) to an existing model architecture such that we have $\{ 
\alpha, \beta, \nu, \gamma
\}$ as the eventual model outputs. 

The mean
prediction, epistemic and aleatoric uncertainties are then defined with respect to $\mathcal{P}$ as follows.  
\be
\label{fNIG2}
\mathbb{E}_{\mathcal{P}}[\mu] = \gamma, \,\,\,\,
U_{a} = \mathbb{E}_{\mathcal{P}}[\text{Var} (f)]=
\mathbb{E}_{\mathcal{P}}[\sigma^2] = \frac{\beta}{\alpha -1}, \,\,\,\,
U_{e} = 
\text{Var}_{\mathcal{P}}[\mathbb{E}(f)] 
=
\text{Var}_{\mathcal{P}}[\mu] = \frac{\beta}{\nu(\alpha -1)},
\ee
where $f$ is the `first-order' PDF of 
eqn.~\eqref{Gaussian}. It is in this sense that
$\mathcal{P}$ is the `higher-order' PDF providing the description of $\mu, \sigma$ of $f$ as random variables. We note that the aleatoric and epistemic uncertainties are obtained after integrating over $\mu, \sigma$
using eqn.~\eqref{fNIG}. The factorized form
of $\mathcal{P}$ enables an analytic expressions for each of these uncertainties. 
In Deep Evidential Learning, the model output parameters are $\{\alpha, \beta, \nu, \gamma \}$. From these parameters, we can then obtain the dose prediction (given by $\gamma$) as well as the uncertainty estimates $U_a, U_e$ for each voxel.

\subsection{Deep Evidential Regression for a dose prediction model: a refined loss function }
\label{sec:loss}

We now apply such an uncertainty quantification framework in the context of a dose prediction model which estimates the dose delivered to each CT voxel based on an input set of CT images and various binary-valued masks representing the radiotherapy's target regions and the organs-at-risk. The OpenKBP dataset of \cite{Babier2,Babier} was used as our source, and a vanilla 3D U-Net was employed as the backbone architecture.

In \cite{amini}, the first-order PDF pertaining to the model output 
$y$ was assumed to be
the normal distribution $\mathcal{N}(y\vert \mu, \sigma )$, 
with a 4-parameter normal-inverse-gamma distribution ( eqn.~\eqref{fNIG} )
adopted to be the
higher-order evidential distribution.
For our case, a crucial caveat is that the radiation dose value is definitively
constrained within a finite interval,
and not unbounded. To incorporate
this physical nature of the radiation dose, 
we first define a dimensionless
dose parameter $y = y(D_p) =  \frac{0.9D_p + 10}{100}$ where $D_p$ is the physical dose in units of Gy. We then pass $y$ to 
its logit representation $\log \left(
\frac{y}{1-y} \right)$ which we take as the form of output for the neural network (rather than the physical dose $D_p$). 
Ablation experiments using the validation dataset were used to determine the coefficients in the linear map
$y = y(D_p)$ such that the logit representation yielded the optimal model performance. Our choice of the linear map $y = y(D_p)$ also avoided the singular points $\{0,1\}$ of the logit function by construction. 

Formally, our choice of the 
output data representation implies that 
the first-order PDF for each voxel is
related to the \emph{logit-normal} distribution more precisely instead of a Gaussian. Its full form reads
\be
\label{logitN}
f \left( y | \mu, \sigma \right) = \frac{1}{\sigma \sqrt{2\pi}}
\frac{1}{y(1-y)}\text{exp}
\left(
-\frac{1}{2\sigma^2}\left[ L(y) - \mu \right]^2
\right), \,\,\,\,\, L(y) \equiv \log  \left( \frac{y}{1-y} \right).
\ee
This is a close cousin of the normal distribution, and has a bounded domain $y\in (0,1)$ as the support. 
The parameters $\mu, \sigma$
are now dimensionless parameters
defined
as the Gaussian mean and standard
deviation of the logit of the (dimensionless) dose $y$. 
For their 
higher-order evidential distributions, 
we can still follow the procedure
of \cite{amini}, i.e. we
adopt a normal distribution for $\mu$ and 
an inverse-gamma distribution for the variance to yield
a normal-inverse-gamma distribution for \eqref{logitN}. Also, similar to the approach in \cite{amini}, we can derive
the aleatoric and epistemic uncertainties. 
In their original forms as defined in 
eqn.~\eqref{fNIG2}, they
are dimensionless variances associated with the logit representation. To convert them to the corresponding dimensionful ones in units of Gy${}^2$, we first invoke
the standard approximation technique of relating the uncertainty in the $L(y)$ (logit of $y$) to that in $y$ itself, before using the linear map $y = y(D_p)$ to deduce the physical aleatoric and epistemic uncertainties in units of Gy${}^2$. 
For clarity, we refer to them separately
as $U_{alea}, U_{epis}$ which are 
related to $U_a, U_e$ in eqn.~\eqref{fNIG2} via
$
U_{alea, epis} = \left( 
\frac{100}{0.9} y(1-y)
\right)^2 U_{a,e}
$.

Following \cite{amini}, 
the primary loss function can be taken to be
the maximum likelihood loss defined
as the marginal likelihood obtained
when we integrate over the first-order mean $\mu$, and variance $\sigma^2$. 
After some algebra, we find that we obtain the likelihood function
\bea
\mathcal{L}_{pri} &=& 
\int^\infty_0 d\sigma^2 \int^\infty_{-\infty}d\mu \,\,
f(y | \mu, \sigma) \, \mathcal{P}(\mu, \sigma | \alpha, \beta, \nu, \gamma ) \cr
\label{maxL}
&=& \frac{1}{y(1-y)} \frac{     
\Gamma( \alpha + \frac{1}{2} )
}{\Gamma(\alpha)} 
\sqrt{\frac{\nu}{\pi}}
\left[
2\beta (1+ \nu)
\right]^\alpha
\left[ 
\nu \left( 
\log \left( \frac{y}{1-y}    \right)
- \gamma \right)^2 + 2\beta (1+\nu)
\right]^{-(\alpha + \frac{1}{2})},
\eea
where the expression lying on the right of the term $1/(y(1-y))$ can be identified
as the Student's \emph{t} distribution $\text{St}  \left( 
\log (\frac{y}{1-y} ) ; \gamma, 
\frac{\beta(1+\nu)}{\nu\alpha}, 2\alpha
\right) $. 
 Unfortunately, we found that 
in this form, the loss function did not lead to stable learning curves that could generate uncertainty heatmaps that correlated with prediction errors. 
After some experimentation, we found that making the following refinements to eqn.~\eqref{maxL} significantly enhanced the effectiveness of the model: 
\begin{itemize}
\item removal of the
factor $1/y(1-y)$ from eqn.~\eqref{maxL},
\item passing the negative log-likelihood
function through a regularizing positive-definite activation function (our choice: standard logistic function),
\item adding a mean-squared-error loss function term.
\end{itemize}
These refinements led to the eventual form of our loss function being as follows. 
\bea
\mathcal{L}_{pri}&=& f_s
\left(
\frac{     
\Gamma( \alpha + \frac{1}{2} )
}{\Gamma(\alpha)} 
\sqrt{\frac{\nu}{\pi}}
\left[
2\beta (1+ \nu)
\right]^\alpha
\left[ 
\nu \left( 
\log \left( \frac{y}{1-y}    \right)
- \gamma \right)^2 + 2\beta (1+\nu)
\right]^{-(\alpha + \frac{1}{2})}
\right),\,\,\,
\cr
\label{loss}
\mathcal{L}_{f} &=&  \mathcal{L}_{pri} + \lambda_{KL} \left|\log \left( 
\frac{y}{1-y}
\right) - \gamma \right| (2\nu + \alpha)
+ \lambda_{mse} \left( 
\log \left( 
\frac{y}{1-y}
\right) - \gamma
\right)^2,\, f_s(g) \equiv \frac{1}{1+g}
\eea
where $\lambda_{KL}, \lambda_{mse}$ are
crucial hyperparameters of which optimal values for our purpose will be discussed in 
Sec.~\ref{sec:modelh}. 
The first regularization term $
\mathcal{L}_{reg} = \left|\log \left( 
\frac{y}{1-y}
\right) - \gamma \right| (2\nu + \alpha)
$ preceded
by $\lambda_{KL}$ was proposed
in \cite{amini} to penalize 
statistical evidence for incorrectly labeled terms.
We note in passing that
in \cite{amini}, it was argued that the prior (the function 
$\mathcal{P}(\mu, \sigma | \alpha, \beta, \nu, \gamma)$ in eqn.~\eqref{fNIG} ) has a singular
limit when we take parameter values
corresponding to `zero evidence', and that
softening this limit by introducing 
finite yet small values for $\alpha = 1+ \epsilon$, 
$\nu = \epsilon$ did not appear to be effective. We found this to be the case as well. Another possibility that we explored is using Jeffrey's prior (which in this case involves an improper uniform prior for the mean of the model output). But this turned out to be similarly ineffective relative to the choice adopted in \cite{amini}.

\subsection{Related Work}
\label{sec:related_work}

Among the many papers devoted to the subject of dose prediction models (see e.g. \cite{Liu} for an extensive review and compilation), to our knowledge, there has been only two which directly delved into the notion of uncertainty quantification frameworks: \cite{Rafe} and \cite{Zhang0}. These papers contained highly interesting results which we would like to briefly discuss in relation to our work.

In \cite{Rafe}, the authors studied 
the frameworks of MC Dropout and bootstrap aggregation which is essentially a Deep Ensemble method where each model is trained on only a portion of the entire training dataset. Like in this work, they used the OpenKBP Challenge dataset of \cite{Babier} for validating the performance of each uncertainty quantification framework, concluding that Deep Ensemble yielded a lower mean absolute error (MAE) while showing better correlation between uncertainty and prediction error. In particular, the authors proposed that with the aid of an additional scaling factor, the models generated uncertainty heatmaps and DVH confidence intervals which appeared reasonable. This scaling factor was defined differently for each 
region of interest (ROI) and was
argued to be necessary to bring the uncertainty values into a more `interpretable scale' as the raw ones were only providing `relative measures'. For each set of points belonging to a ROI in the validation dataset, this ROI-dependent factor was defined simply as the standard deviation of the ratio between the prediction error and the uncertainty value 
(see eqn.~(3),(4) of \cite{Rafe}). 
In our work, we did not incorporate such an empirical scaling factor in our definitions of various uncertainties, including our final uncertainty heatmaps and the DVH confidence intervals. For Deep Evidential Learning where the predictive variance can be expressed precisely in terms of learned output variables of the model, in principle, it is not clear to us how an additional empirical scaling factor should be justified. By construction, this rescaling factor naturally enhances the correlation between uncertainty and error since the uncertainty values themselves are part of the factor's definition, assuming that it generalizes well for unknown (e.g. testing) data. Conversely, we took the resulting average value of such a ratio (across all voxels) to assess whether the uncertainty distribution generated by the model was reasonable in terms of the overall scale of magnitude. Through the numerical values (see Table \ref{table:1}) and the DVH certainty intervals in Fig.~\ref{fig:DVH_patient}, we found that this was indeed the case.

The work in \cite{Zhang0} proposes 
the use of Gaussian mixture models of which parameters are the neural network's outputs, and with U-Net as the backbone architecture. Such a mixture density network is very similar in form to our Deep Evidential Learning model since the output variables of our model are also parameters of a PDF. In our case, the PDF is a higher-order normal-inverse-gamma distribution whereas in 
\cite{Zhang0}, the parameters are the means and variances of the component
normal distributions (see eqn.~\eqref{Gaussian}) and the coefficients defining their linear combination. Each component of the Gaussian mixture pertained to a treatment protocol representing specific priorities (e.g. OAR sparing over target coverage, etc.), and the relative values of the variances quantified the trade-offs between these protocols. 
The cross-entropy loss function associated with the cumulative distribution function (CDF) of the predicted dose was then taken as part of the objective function for generating a dose mimicking model. In \cite{Zhang0}, deliverable treatment plans were the endpoints of their dose prediction algorithm, with the optimization problem solved by RayStation's native sequential 
quadratic programming solver and final dose computed via a collapsed cone algorithm \cite{Zhang0}. Although we noted that there was no scrutiny of the correlation between uncertainty and error, the authors of \cite{Zhang0} (and its follow-up work \cite{Zhang}) notably pointed out that probabilistic dose prediction models such as \cite{Rafe} and ours could play a crucial role in their pipeline since our model would yield dose
CDFs that can be used to define the cross-entropy loss function as part of the dose-mimicking model's objective function. It would thus be interesting to extend results of our work here by integrating our probabilistic dose predictions into a pipeline like \cite{Zhang0} where deliverable treatment plans with specified machine parameter settings can be anticipated as final outcomes.

\section{Methodology}
\label{sec:M}
\subsection{On the OpenKBP Challenge Dataset}
The OpenKBP Challenge dataset of \cite{Babier2, Babier}
is devoted towards establishing an open framework for the development of plan optimization models for knowledge-based
planning (KBP) in radiotherapy. 
It is an augmented variation
of real clinical data constructed as follows: medical images were taken from several institutions available on 
The Cancer Imaging Archive (TCIA)
\cite{TCIA} which hosts open-source data that has undergone de-identification compliant with DICOM standard.\footnote{In particular, all metadata and Protected Health Information (PHI) has been removed from the datasets provided in OpenKBP Challenge \cite{Babier2,Babier}.} After removing highly incomplete imaging datasets to obtain a final competition dataset comprising of 340 patients, synthetic radiation plans were generated for each of them using a published automated planning method \cite{Aaron}. 
This final dataset is then split randomly by the organizers of OpenKBP Challenge to be as follows: 200 (training), 40 (validation), 100 (testing). In our work, we followed this decomposition of data. As in a conventional machine learning pipeline, the validation dataset was used for tuning of hyperparameters whereas the testing dataset was used for reporting various results in our work. We refer the reader to \cite{dataset, validation} for more pedagogical expositions of data-splitting practices and algorithms.

The radiotherapy plans were delivered using 
nine equidistant coplanar beams at various angles with a 6 MV step-and-shoot
intensity-modulated-radiation-therapy (IMRT) 
in 35 fractions. 
The organizational team of the Challenge used the IMRTP 
library from \emph{A Computational Environment for Radiotherapy Research} implemented in MATLAB \cite{Deasy} to generate
the dose deposited at each voxel. These dose distributions were from fluence-based treatment plans with similar degrees of complexity
\cite{Babier,Craft2}. 

We used a $(128, 128, 128, 11)$-dimensional
input representation consisting of 
(i)CT in Hounsfield units(clipped to be within [0, 4095] before
being normalized to [0,1]), 
(ii)structure masks of three planning target volumes (PTV) and seven organs-at-risk (OAR) regions : each is a Boolean tensor 
labeling any voxel contained within the respective structure. The OARs are brainstem, spinal cord, right and left parotids, larynx,
mandible and esophagus. The three PTV regions are: PTV56, PTV63, PTV70 which are targets that should receive 56 Gy, 63 Gy and 70 Gy of radiation dose respectively.

\subsection{On model architecture and hyperparameters}
\label{sec:modelh} 
Since our primary goal is to examine 
the uncertainty estimation aspects of the model, in this work, we adopt a simple vanilla 3D U-Net as the backbone architecture upon which we insert two additional layers so that they are compatible with the framework of Deep Evidential Learning. 
U-Nets and their variants have featured heavily in many deep-learning-guided tasks related to radiotherapy dose prediction \cite{Rafe, Babier}. They are essentially convolutional neural networks equipped with an encoder-decoder architecture that allows effective learning of both image features at various levels of resolution (see e.g. \cite{UReview,unet2} for reviews on U-Net and \cite{conv1,conv2} for pedagogical introductions to convolutional neural networks).

\begin{figure}[h!]
\centering
\includegraphics[width=0.9\textwidth]{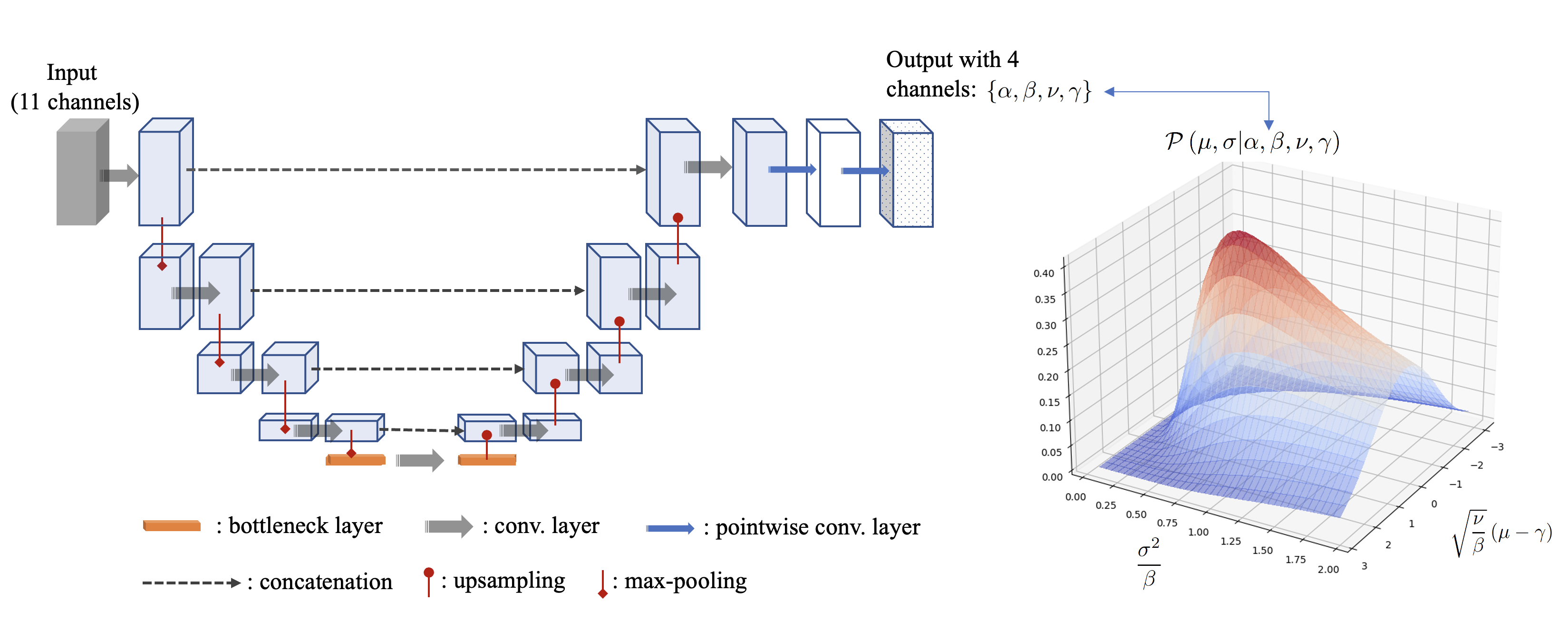}
\caption{A sketch of the Deep Evidential model with 3D U-Net backbone architecture.
The 4-channel outputs of the model are the parameters of a
normal-inverse-gamma distribution schematically plotted on the right.  
Details of the convolutional and other layer operations in the backbone segment are described earlier in Sec.~\ref{sec:modelh} and largely identical to the original 3D U-Net of \cite{olaf}. We adjoined the U-Net structure to the Deep Evidential framework by passing the output obtained after the final upsampling layer to two consecutive pointwise convolution layers with number of channels = 8, 4 respectively. 
The final output has dimensions
$(128, 128, 128, 4)$.}
\label{fig:ModelS}
\end{figure}

Our U-Net has the following structure (see Fig. \ref{fig:ModelS}): 
\begin{itemize}
\item it takes in a 11-channel input of size
$128 \times 128 \times 128$ voxels,
\item each downsampling level consists of
two convolutional layers each with a $(3 \times 3 \times 3)$ kernel, ReLU activation and 
equipped with a dropout unit, followed by maximum pooling with kernel size $(2\times 2 \times 2)$,
\item the downsampling operation is performed 4 times, with the dropout rates increasing consecutively as $\{0.10, 0.15, 0.20, 0.25\}$, and
the number of convolutional filters for each layer being $\{16, 32, 64, 128\}$ respectively, 
\item the bottleneck layer has 256 filters and
dropout rate of 0.30,
\item at each of the four upsampling levels, features from the contraction path are concatenated with the corresponding upsampled features, and the dropout rates decrease similarly in the reverse order. 
\item a 8-channel pointwise convolution is then applied followed by a final 4-channel pointwise convolution that yields the four parameters 
$\{\alpha, \beta, \nu, \gamma \}$ of the Bayesian prior distribution $\mathcal{P}(\mu, \sigma| \alpha, \beta, \nu, \gamma)$ of eqn. \eqref{fNIG}.
\end{itemize}
This model carries about $6\times 10^6$ free weight parameters.
Various hyperparameters such as the dropout rate at each level, etc. were eventually adopted after running ablation experiments to determine their optimal values. 

Imposing some choice of numerical bounds to 
$\{\alpha, \beta, \nu, \gamma \}$ in the final layer is a crucial aspect of the hyperparameter tuning process unique to this framework. The parameter $\beta$ sets the overall (dimensionless) scale to both uncertainties of which ratio is set by $\nu = U_{alea}/U_{epis}$ which has to be positive. 
For this dataset, we found good uncertainty-error correlation after imposing $\beta \geq 10^{-3}$. In \cite{amini}, the authors imposed the constraint $\alpha > 1$ which we also followed here, as for this range of values, the aleatoric and epistemic uncertainties can be conveniently described by simple analytic expressions. More broadly speaking,
$\alpha >0 $ is the larger admissible range for $\alpha$ as a parameter of the inverse-gamma distribution. We also found it useful to 
map the zero dose to a small positive value 
$\epsilon = 0.1$ in the (dimensionless) normalized dose $D_N$ defined to be $D_N = \frac{0.9 D_p + 10}{100}$ where $D_p$ is the physical dose in Gy. In the loss function, we work in the logit-representation of $D_N$,
with the predicted dose $\gamma = \log\left(
\frac{D_N}{1-D_N}
\right)$. Since this logistic function is monotonic, $\gamma$ is bounded from below as
$\gamma \geq \log \left(   \frac{\epsilon}{1- \epsilon }  \right) = -\log 9 $ with our choice of $\epsilon = 0.1$.

\begin{figure}[h!]
 \centering
 \includegraphics[width=0.9\textwidth]{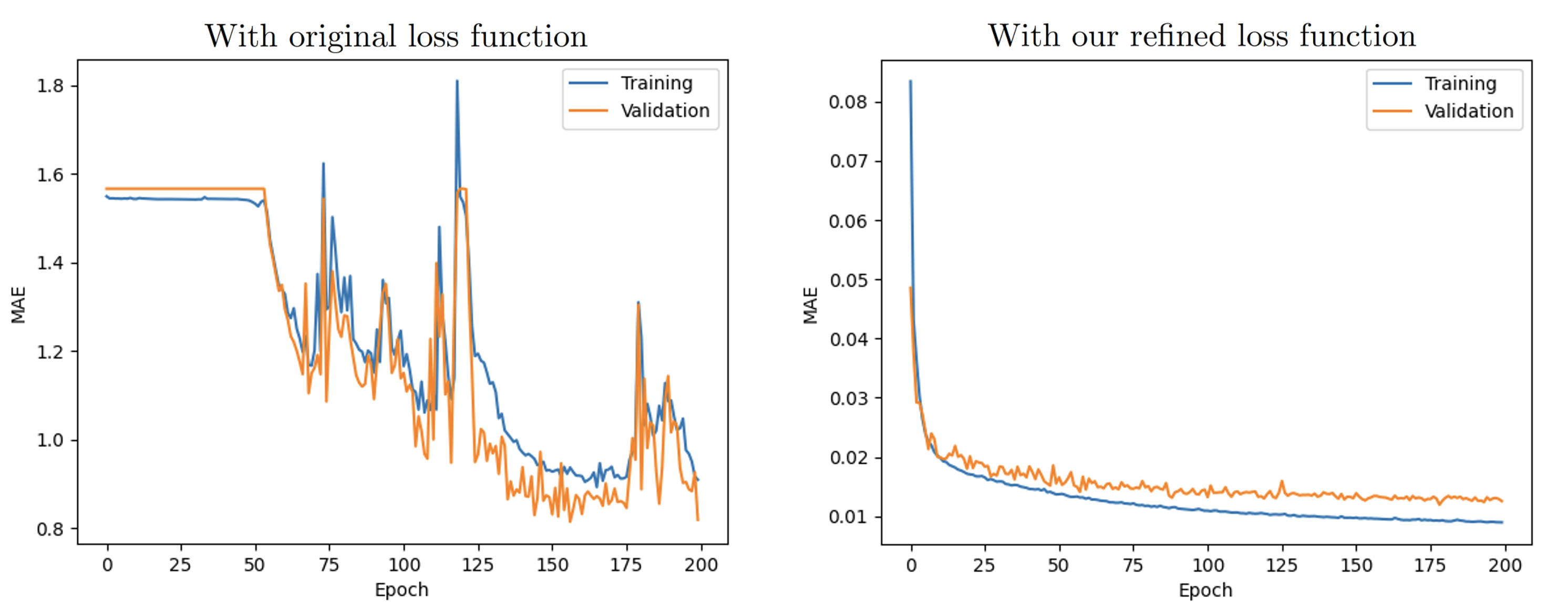}
 \caption{Plots of the validation and training mean-absolute-error (MAE) for the model equipped with the original (left) and our refined (right) loss functions.     }
\label{fig:lossC}
\end{figure}

For the loss function originally proposed 
in \cite{amini}, we found that unfortunately, it
led to training curves with frequent oscillations and poor eventual outcomes. As explained in detail in Sec.~\ref{sec:loss}, we reformulated the
loss function for a logistic representation of the dose variable such that one could obtain a stable implementation. 
Our refined loss function in eqn. \eqref{loss} is
characterized by the hyperparameters $\lambda_{mse}, 
\lambda_{KL}$ which describe couplings of the loss function to 
a mean-squared-error (MSE) term and a KL-divergence-like regularization term respectively. Results from ablation experiments with validation dataset indicated that optimal values of these 
hyperparameters lie within the intervals $\lambda_{mse} \in (0.01, 0.1)$, $\lambda_{KL} \lesssim 0.01$. For the results reported here for the rest of our paper, we took $\lambda_{mse} = 0.05,\, \lambda_{KL} = 0.01$. In Fig. \ref{fig:lossC}, 
we plot the learning curves pertaining to the original and our refined loss functions for comparison.

Towards comparing the Deep Evidential model against more common approaches of uncertainty quantification in literature, we implemented a Deep Ensemble model and Monte-Carlo Dropout model for the same dataset. 
Each of the 5 neural networks used in the ensemble model shared the same 3D U-Net architecture with all hyperparameters preserved. Random weight initialization was all performed via the \emph{He}-uniform initializer \cite{He}. Similarly, for the
Monte-Carlo Dropout model, the same base model
was used for passing 30 forward passes with the dropout layers activated for model prediction. 
All models were trained for 200 epochs with a learning rate of $10^{-4}$ using Adam optimizer. 
Our focus in this work lies in studying
aspects of uncertainty estimation, and thus
our choice of a relatively simple backbone architecture equipped with only $\sim 6 \times 10^6$
weight parameters. Compared to other more complicated
setups, each of these models converged quickly in about a couple of hours while attaining a mean-absolute-error of dose prediction that laid within the top 20 of the score chart of the OpenKBP Challenge \cite{Babier}.

\section{Results}
\label{sec:Results}

We found that our Deep Evidential Learning model
yielded uncertainty estimates which demonstrated
strong correlation with prediction errors, while achieving a similar level of accuracy relative to the methods of Monte-Carlo Dropout and Deep Ensemble. The metrics that we used as measures of any putative uncertainty-error associations were (i)the Spearman's rank correlation coefficient between the uncertainty and error values across the testing dataset and their patient-averaged
distributions, (ii)the mutual information between them.

\begin{table}[h!]
\begin{center}
\begin{tabular}{ |c||c|c|c|c|c| } 
\hline
\hline
$\,$ & $\,$ & $\,$ & $\,$ & $\,$ & \\[-1em]
 \textbf{Model} & MAE (Gy) &
 $U_{avg}$ (Gy${}^2$) &
 $r_s(\overline{U}, \overline{D_e})$ & 
 $r_s(U, D_e)$ &
 $M.I. (U, D_e) $
 \\ 
 $\,$ & $\,$ & $\,$ & $\,$ & $\,$ & \\[-1em]
\hline
 Deep Evidential $(U_e, U_a)$ & 
 3.09 & (2.80, 0.20) & (0.83, 0.69) & 
 (0.69, 0.63) & (0.67, 0.61) \\ 
 Monte-Carlo Dropout & 3.28 & 2.12 & 0.73 & 0.62 & 0.60 \\ 
Deep Ensemble & 3.10 & 3.42 & 0.86 & 0.66 & 0.56 \\
\hline
\hline
\end{tabular}
\caption{Table collecting 
various measures of uncertainty-error
correlations. \emph{Abbreviations}:
MAE = mean absolute error,
$U_{avg}$ = mean uncertainty value, 
$r_s(\overline{U}, \overline{D_e})$
= Spearman's coefficient between patient-averaged uncertainty and error 
distributions, 
$r_s (U, D_e)$ = Spearman's coefficient between uncertainty and error distributions, $M.I.(U, D_e)$ = 
mutual information between uncertainty and error distributions, $D_e$ = prediction error, $(U_e, U_a)$ = (epistemic uncertainty, aleatoric uncertainty).
}
\label{table:1}
\end{center}
\end{table}
As summarized in Table
\ref{table:1}, one observes that across the 
various correlation indices, the 
epistemic uncertainty's degree of association with error was higher than 
Monte-Carlo Dropout and Deep Ensemble, with the only exception being its
$r_s(\overline{U}, \overline{D_e})$
slightly lower than that of Deep Ensemble. 
For all three models, 
the correlation between patient-averaged
uncertainties and errors was always higher, suggesting that there is a certain degree
of stochasticity characterizing this relationship within each patient's dataset. 
Relative to epistemic uncertainty $U_e$, aleatoric uncertainty $U_a$ in the Deep Evidential Learning model consistently demonstrated lower correlations with error, and its $r_s$ value was less affected by averaging within individual patient (compared to other models). 
We also note that the order-of-magnitudes
of the various averaged uncertainty values
$U_{avg}$ were such
that $\text{MAE}/\sqrt{U_{avg}} \sim 2$, the exception being $U_a$ which was generally an order-of-magnitude below $U_e$ and uncertainties of the other two models.

\subsection{Variation of prediction error with uncertainty threshold}

\begin{figure}[h!]
 \centering
 \includegraphics[width=\textwidth]{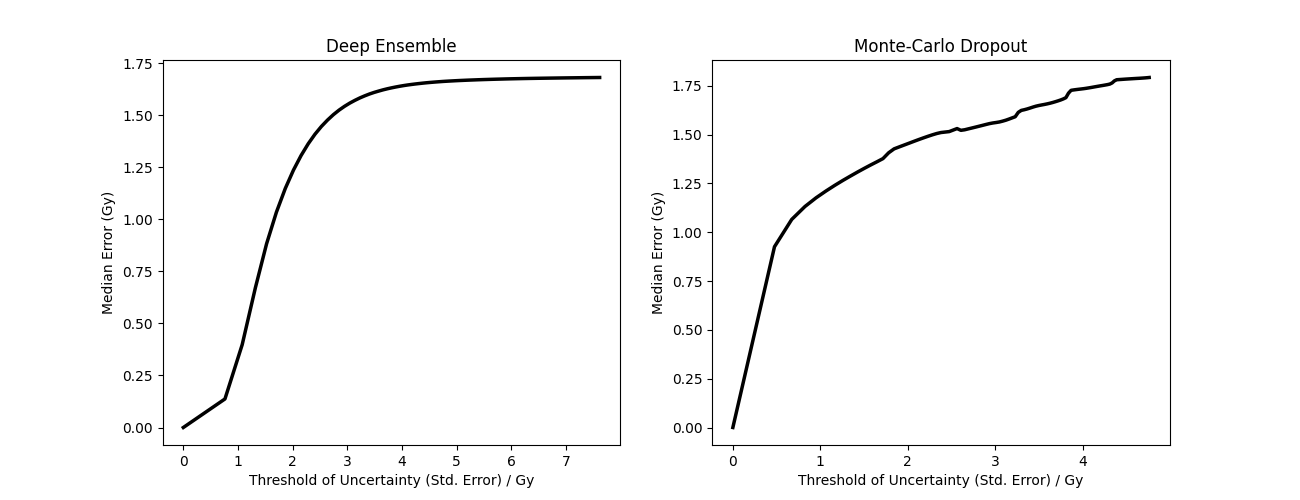}
 \caption{The median error plotted is that of the subset of the testing dataset which remains after imposing an upper threshold value for the uncertainty. We note that each model generated different ranges of uncertainty values.    }
\label{fig:DE1}
\end{figure}

\begin{figure}[h!]
 \centering
 \includegraphics[width=\textwidth]{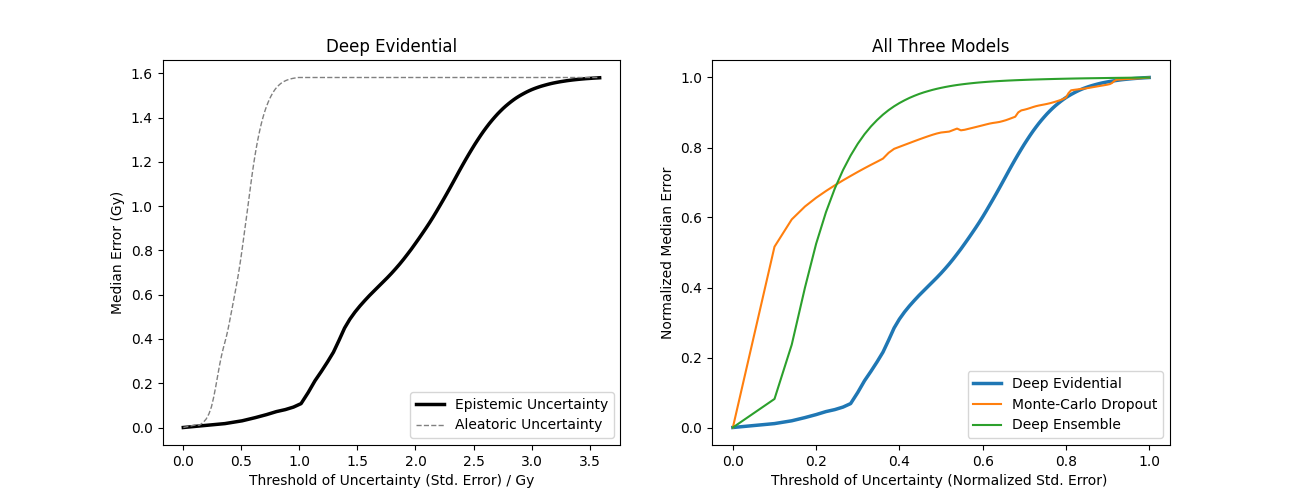}
 \caption{The left diagram pertained to Deep Evidential Learning where we included both the curves for aleatoric and epistemic uncertainties. On the right, we furnished a comparative plot where both median error and uncertainty threshold were normalized using their maximum values. This allows one to differentiate among the curves based on a common overall scaling.   }
\label{fig:DE2}
\end{figure}

The results in Table 
\ref{table:1} indicated that Deep Evidential Learning model yielded spectra of uncertainty values which were more correlated with prediction error, relative to MC Dropout and Deep Ensemble methods. 
To further scrutinize the difference
among the various frameworks, we 
examined how error distributions varied
with changing threshold values of uncertainty. 
Since the error distributions were generically found to be quite skewed, we used the median of the error distribution as its characterizing parameter.  
Fig.~\ref{fig:DE1} and \ref{fig:DE2} below reveal how median error changed with increasing levels
of uncertainty thresholds for the different models. 
Assuming that uncertainty values correlate positively with prediction error implies that one expects a monotonically increasing curve in such a plot, a trend that was indeed manifest for these models
as shown in Fig.~\ref{fig:DE1} and \ref{fig:DE2}. 

In particular, 
from Fig.~\ref{fig:DE2},
the visibly more evident linearity of the curve for Deep Evidential Learning indicated a sensitivity of epistemic uncertainty to prediction errors that was relatively more uniformly calibrated compared to Monte-Carlo Dropout and Deep Ensemble methods.

\subsection{Probing noise-sensitivity of uncertainty measures}

A prominent feature of Deep
Evidential Learning is a clear 
principled approach towards
distinguishing between aleatoric and epistemic uncertainties. They can be defined precisely using the higher-order PDF $\mathcal{P}$ as expressed in eqn.~\eqref{fNIG2}. 
Traditionally, aleatoric uncertainty is commonly interpreted as arising from inherent random noise in the data, in contrast to epistemic uncertainty which presumably captures uncertainty originating from model suitability and data sufficiency (see e.g. \cite{zouReview,kendall}). 

We would like to probe the level of sensitivity of various uncertainty measures to noise perturbations of the input data. In particular, we would like to identify if there would be any difference in noise-sensitivity between $U_{alea}$ and $U_{epis}$. 
Our choice of perturbation is the addition of Gaussian noise \cite{gravel} of mean zero and standard deviation 0.5 to each CT voxel (of which intensity was preprocessed to lie within $[0,1]$). 
To enable a more holistic description of response to noise, we plot the empirical cumulative
distribution function (eCDF) of various uncertainty measures preceding and following the noise addition. 

\begin{figure}[h!]
  \centering
  \includegraphics[width=0.8\linewidth]{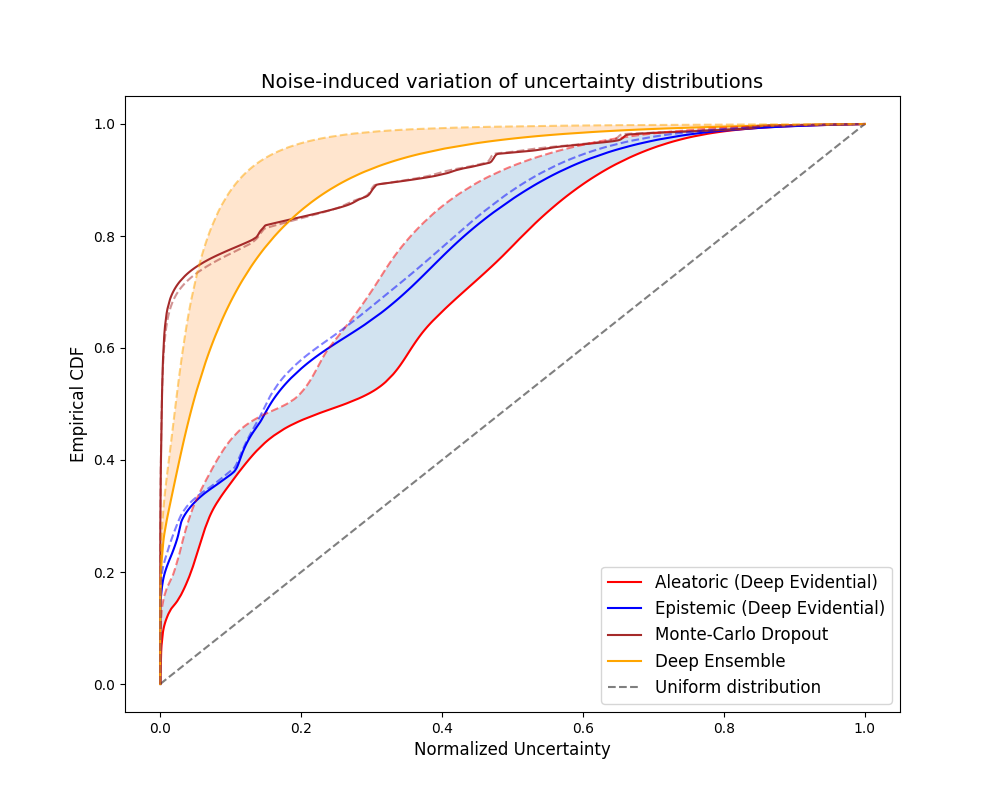}
\caption{Dashed lines are the uncertainty 
eCDF in the absence of noise, while
the corresponding solid curves
show distributions in the presence of 
an added Gaussian noise with zero mean
and standard deviation of 0.5 for the
CT intensity value in each voxel. 
Shaded regions accentuate the more significant shift associated with aleatoric uncertainty (of the Deep Evidential Model) and Deep Ensemble's uncertainty distributions.}
\label{fig:NoisePic}
\end{figure}

In Fig.~\ref{fig:NoisePic}, 
the plot of various eCDFs collectively illustrated
the differences among the uncertainty measures in their responses towards the Gaussian perturbation. Since different measures yielded distinct overall scale of uncertainty magnitudes, we normalized each with respect to the maximum value for each uncertainty type. In the limit of the Gaussian noise completely dominating over any pre-existing noisiness, and assuming
that the uncertainty value of each voxel is correlated only with the noise PDF (identical for every voxel), we would expect the uncertainty eCDF to approach a symmetric distribution with mean (normalized) uncertainty at 0.5 in this limiting scenario. As a simple reference distribution, in Fig.~\ref{fig:NoisePic}, we included the line corresponding to the uniform distribution.
From just visual inspection, 
aleatoric uncertainty distribution and that of Deep Ensemble changed 
more significantly relative to others. The noise-induced fractional changes in the relative entropy (measuring the Kullback-Leibler divergence between each eCDF and the uniform distribution) turned out to be $\sim 0.1$ for aleatoric uncertainty and Deep Ensemble, about ten times larger than those for epistemic uncertainty and Monte-Carlo dropout.

The various curves appeared to suggest the following. 
\begin{itemize}
\item The relative responsiveness of aleatoric (stronger) and epistemic (weaker) uncertainties 
are indeed compatible with their conventional interpretations as reflecting inherent data noise and 
model-related uncertainties respectively.
\item Monte-Carlo Dropout yielded an uncertainty distribution that, like epistemic uncertainty, was not sensitive to the addition of the Gaussian noise. This favored its interpretation as an epistemic uncertainty in nature. 
\item Deep Ensemble method yielded an uncertainty distribution that was responsive to the addition of noise. Like the aleatoric uncertainty in Deep Evidential Learning, the eCDF shifted to being 
less long-tailed and leaning towards being more uniform. 
\end{itemize}
For the Deep Ensemble method, it was explained
in \cite{lak,Toro} that one can often use a Gaussian mixture model as an effective description of the ensemble prediction. By Eve's law of total variance \cite{Toro}, the predictive variance admits a natural decomposition into aleatoric and epistemic components. This appears to be consistent with the noise sensitivity displayed by Deep Ensemble method in Fig.~\ref{fig:NoisePic}.

\subsection{Applications}

\subsubsection{Uncertainty heatmaps}
Heatmaps of the dose uncertainty distributions can 
be used through direct visual inspection to discover regions of potentially acute errors made by the dose prediction model. This capability hinges
on the strength of the correlation between uncertainty and error distributions. In Table \ref{table:1}, 
epistemic uncertainty in the Deep Evidential model appeared to perform comparatively well, if not better, than the two more conventional
uncertainty quantification frameworks as measured by mutual information and Spearman's correlations. 
These statistical indices were defined at the level of the entire (testing) dataset. For each of the seven OAR and three PTV regions, one could in principle compute the uncertainty distribution localized within its interior. Thus, in Table \ref{table:2} we collect for various ROIs their mean uncertainty ($U_{alea}, U_{epis}$) values together with the errors in predicted dose $D_e$.  The Spearman's coefficients between each uncertainty type and the corresponding dose prediction error $\overline{D}_e$ are also indicated (all with p-values $p<.01$). Among all the ROIs, the L,R-parotid regions exhibited the highest correlations between $\overline{D}_e$ and epistemic uncertainty, while the larynx region showed the lowest measures of correlation for both types of uncertainties. The mandible region carried the highest aleatoric uncertainty and was associated with the highest $\overline{D}_e$. 
Across the spectrum of ROIs in Table \ref{table:2}, we found each mean epistemic uncertainty to be strongly correlated with $\overline{D}_e)$, with Spearman's correlation coefficient of $0.80\,(p=0.005)$, whereas that for the aleatoric uncertainty was much less significant at $0.57\, (p=0.09)$.

\begin{table}
\begin{center}
\begin{tabular}{ |c||c|c|c|c|c| } 
\hline
\hline
$\,$ & $\,$ & $\,$ & $\,$ & $\,$ & \\[-1em]
 \textbf{ROI} & $\overline{U}_{alea}$ (Gy${}^2$) & $\overline{U}_{epis}$ (Gy${}^2$)  & $r_s(U_{alea}, D_e)$ &
 $r_s(U_{epis}, D_e)$ &
 $\overline{D}_e $ (Gy)
  \\ 
  \hline
 PTV70 & 0.21 & 2.63 & 0.56 & 0.68 & 2.23 \\ 
 PTV63 & 0.21 & 2.65 & 0.65 & 0.74 & 2.16 \\ 
 PTV56 & 0.20 & 2.36 & 0.68 & 0.78 & 1.87 \\
 Brainstem & 0.10 & 0.94 & 0.69 & 0.69 & 0.65 \\
 Spinal Cord & 0.27 & 2.49 & 0.79 & 0.82 & 1.59 \\
 R-Parotid & 0.13 & 2.43 & 0.80 & 0.84 & 1.38 \\
 L-Parotid & 0.10 & 2.78 & 0.80 & 0.84 & 1.45\\
 Esophagus & 0.06 & 1.18 & 0.71 & 0.71 & 0.98 \\
 Larynx & 0.17 & 1.95 & 0.43 & 0.48 & 2.46 \\
 Mandible & 0.21 & 3.77 & 0.83 & 0.79 & 4.02\\
 \hline
 \hline
\end{tabular}
\caption{Table collecting various ROIs, their associated uncertainty ($U_{alea}, U_{epis}$) values together with the errors in predicted dose $D_e = | D_p - D_{GT} | $. These values were averaged over the entire testing dataset. $r_s(U_{alea}, D_e)$ and $r_s(U_{epis}, D_e)$ denote the Spearman's coefficients between each uncertainty type and the corresponding dose prediction error $D_e$. 
}
\label{table:2}
\end{center}
\end{table}

At the level of each ROI, apart from the larynx, $r_s (U_{epis}, D_e) \gtrapprox 0.7$ (1 s.f.) for all other OAR and PTV regions. 
Coupled with the error-uncertainty threshold study in Sec.~4.1 and the noise-sensitivity test in Sec.~4.2, this tapestry of results equips us with the basis for interpreting elevated heatmap regions of epistemic uncertainty as indicators of potential regions of prediction errors. 

\begin{figure}[h!]
  \centering
  \includegraphics[width=\linewidth]{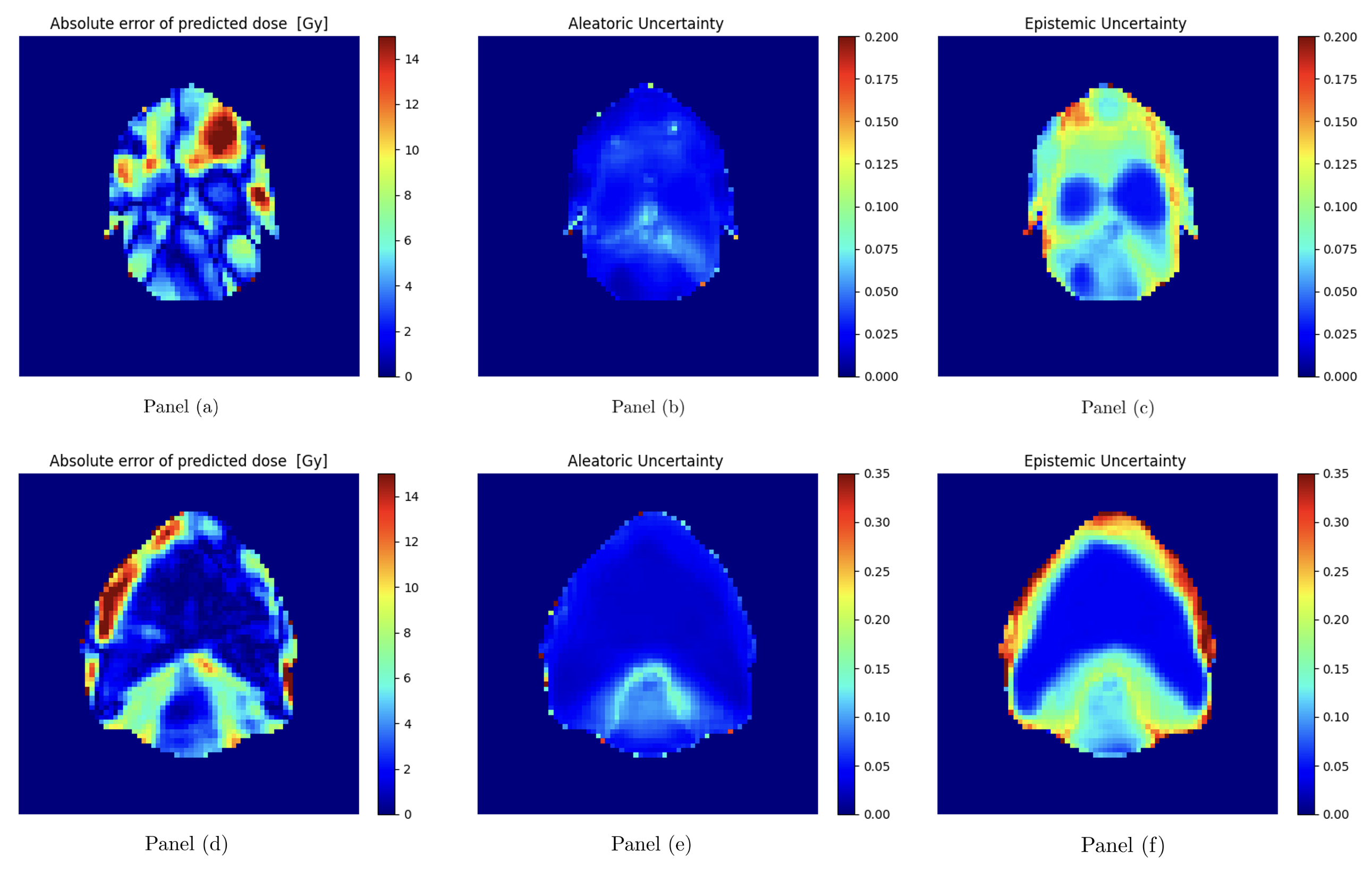}
\caption{Axial CT slices depicting model prediction errors and their uncertainty heatmaps. Each row corresponds to an example patient in the testing dataset of \cite{Babier}; panels (a) and (d) show the mean absolute dose prediction error heatmap; panels (b) and (e) show the aleatoric uncertainty distribution,
while panels (c) and (f) depict the epistemic uncertainty distribution (normalized within each image).}
\label{fig:DosePic}
\end{figure}

In Fig.~\ref{fig:DosePic}, we display an illustrative set of axial CT images portrayed alongside their corresponding uncertainty heatmaps. In principle, the aleatoric uncertainty probes the level of inherent data noise, while the epistemic uncertainty is more strongly correlated
with the elevated regions in the (leftmost) dose prediction error heatmap. Indeed, in Fig.~\ref{fig:DosePic}, we observe that the relative intensity map of epistemic uncertainty tends to parallel the corresponding error heatmap more sensitively. 

\subsubsection{DVH with confidence bands}
\label{sec:DVHbands}
For each patient, we can combine the aleatoric and epistemic uncertainties learned by the neural network to construct confidence intervals for the Dose-Volume-Histogram for each of the region-of-interest. 
Denoting the predictive variance of the dose in each voxel by $\delta D^2_p$, and the predicted dose variable by $\mu_p$, one can invoke the general Eve's law of total variance to obtain
\be
\label{Eve0}
\delta D^2_p \equiv
\text{Var}[\mu_p] = \mathbb{E}\left[
\sigma^2
\right]+ \text{Var} \left[ \mu \right],
\ee
where $\mu, \sigma$ are the mean and variance defined
in \eqref{mu_sigma}. 
To see this, we recall that in general, 
Eve's law of total variance (see e.g. \cite{weiss}) states the following relation for random variables $Y$ and $X$:
\be
\label{Eve1}
\text{Var}[Y] = \mathbb{E}\left[
\text{Var}\left[ Y | X \right]
\right]+ \text{Var} \left[ 
\mathbb{E}\left[ Y | X \right]
\right],
\ee
where conditional expectations such as $\mathbb{E}\left[ Y | X \right]$ are random variables themselves. To apply
 \eqref{Eve1} appropriately to our context, 
we can identify $Y \sim \mu_p$, and $X$ to collectively represent all model-dependent and data-dependent variables. The relation \eqref{Eve0} then follows directly
from \eqref{Eve1} by virtue of the definitions of $\sigma^2, \mu$. In the following, we furnish a contextual proof of this result taking into account our framework explicitly.\footnote{Our style of proof is similar in spirit to that presented in \cite{Toro} for a class of Deep Ensemble models that admit interpretations as Gaussian mixture models, and of which outputs are engineered to be $\mu, \sigma$.} We begin with the definition
\be
\label{Eve3}
\text{Var} \left[ \mu_p \right] = \mathbb{E} \left[
\mu^2_p
\right] - \overline{\mu}^2_p, \,\,\,
\overline{\mu}_p \equiv \iint d\mu d\sigma \,\, 
\mathcal{P}(\mu, \sigma | \vec{\alpha} )\,\, \mu, \,\,\,\,\,
\vec{\alpha} \equiv \{ \alpha, \beta, \nu, \gamma \}.
\ee
Recall that in the Deep Evidential Learning framework, a higher-order
prior distribution $\mathcal{P}$ describes the randomness of $\sigma, \mu$. Also, taking into account the definition of $\sigma, \mu$, we can then write 
\be
\label{standardD}
\iint d\mu d\sigma \, \mathcal{P}(\mu, \sigma | \vec{\alpha} )\, \sigma^2 
= 
\iint d\mu d\sigma \, \mathcal{P}(\mu, \sigma | \vec{\alpha} )\,\left( \mu^2_p - \mu^2 \right).
\ee
Substituting \eqref{standardD} into \eqref{Eve3}, we then have
\bea
\delta D^2_p \equiv \text{Var} [\mu_p ] &=& \iint d\mu d\sigma \, \mathcal{P}(\mu, \sigma | \vec{\alpha} )\, \left( 
\sigma^2 + \mu^2
\right) - \overline{\mu}^2_p \cr
&=& \mathbb{E}[\sigma^2 ] + 
\iint d\mu d\sigma \, \mathcal{P}(\mu, \sigma | \vec{\alpha} )\, \left( \mu^2 - \overline{\mu}^2_p \right) 
= \mathbb{E} [\sigma^2] + \text{Var}[\mu ],
\eea
hence recovering eqn.~\eqref{Eve0} expected from more general arguments. Thus, we see that 
the predictive variance is the sum of the
aleatoric and epistemic uncertainties. 
Our derivation is consistent with a similar result expressed in \cite{kendall}, and provides the theoretical basis for an operational definition of $\delta D_p$ from knowledge of aleatoric and epistemic uncertainties. 

\begin{figure}[h!]
    \centering
    \begin{subfigure}[b]{0.40\textwidth}
        \centering \includegraphics[width=\textwidth]{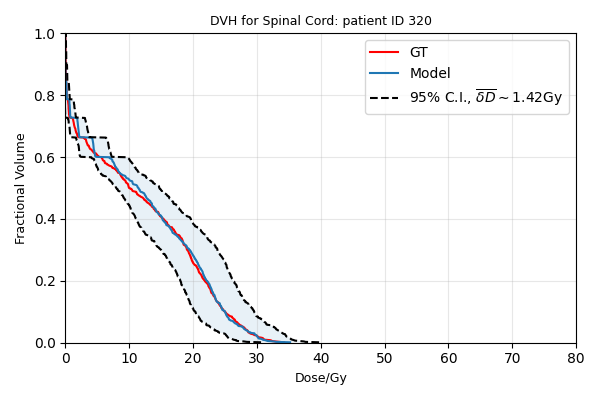}
        \caption{$\,$}
        \label{fig:sub1}
    \end{subfigure}
    \begin{subfigure}[b]{0.40\textwidth}
        \centering       \includegraphics[width=\textwidth]{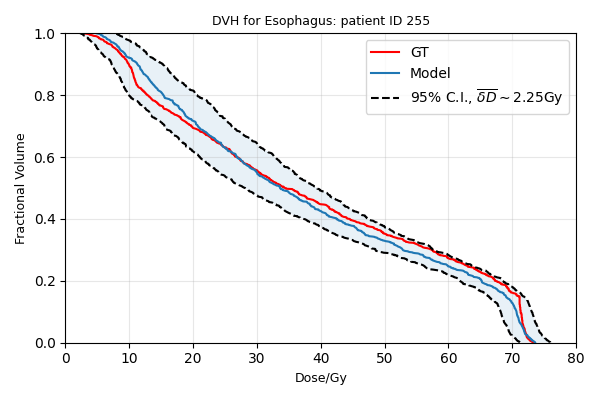}
        \caption{$\,$}
        \label{fig:sub2}
    \end{subfigure}
    \begin{subfigure}[b]{0.40\textwidth}
        \centering     \includegraphics[width=\textwidth]{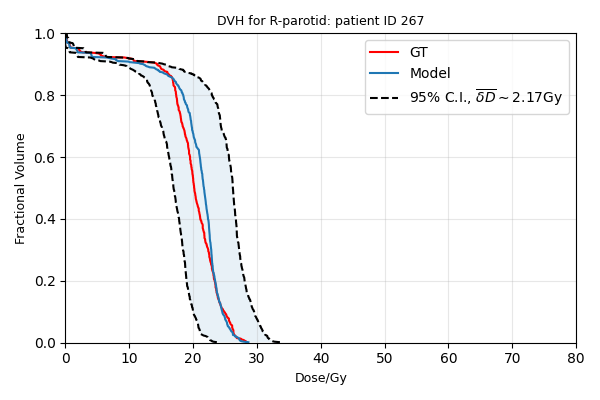}
        \caption{$\,$}
        \label{fig:sub3}
    \end{subfigure}
    \begin{subfigure}[b]{0.40\textwidth}
        \centering
    \includegraphics[width=\textwidth]{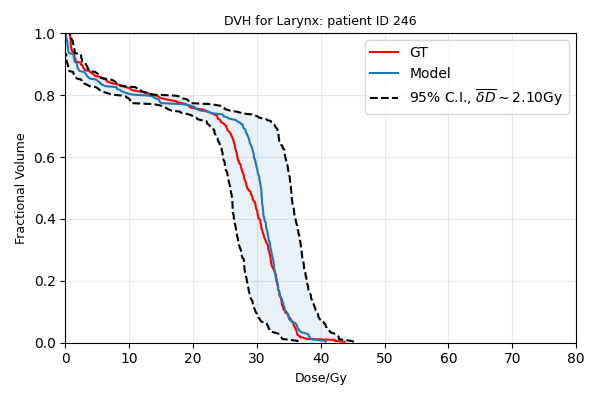}
        \caption{$\,$}
        \label{fig:sub4}
    \end{subfigure}
    \caption{A set of examples of DVH for spinal cord, esophagus, right parotid and larynx each equipped with confidence bands. Dashed curves are the bounds for the $95\%$ intervals defined by $\pm 1.96 \, \delta D_p \equiv \pm 1.96 \sqrt{U^2_{alea} + U^2_{epis}}$. Red and blue curves pertain to the ground truth and model prediction respectively.  }
    \label{fig:dvhA}
\end{figure}

In Fig.~\ref{fig:dvhA}, we plot examples of individual patient's DVH for various PTV and OAR regions, each equipped with a $95\%$ confidence interval. In each diagram of Fig.~\ref{fig:dvhA}, $\overline{\delta D}$ refers to the standard error in dose in units of Gy obtained by averaging $\delta D_p$ over all voxels of the ROI in the patient whose ID shown is defined in the dataset of \cite{Babier}.
A visible feature of these bands is that they tend to contain the more extreme deviations of each groundtruth DVH from its corresponding predicted one along their peripheral edges. 
This is consistent with the order-of-magnitude estimate of $MAE/\delta D_p \sim 1.7$ as drawn from numerical values of the mean error and uncertainties in Table~\ref{table:1}. Thickness and various features of these DVH confidence bands for the OARs and PTV regions can be used to characterize the extent of reliability of various dose prediction models, differentiating among them in this aspect.

\begin{figure}[h!]
  \centering
  \includegraphics[width=0.8\linewidth]{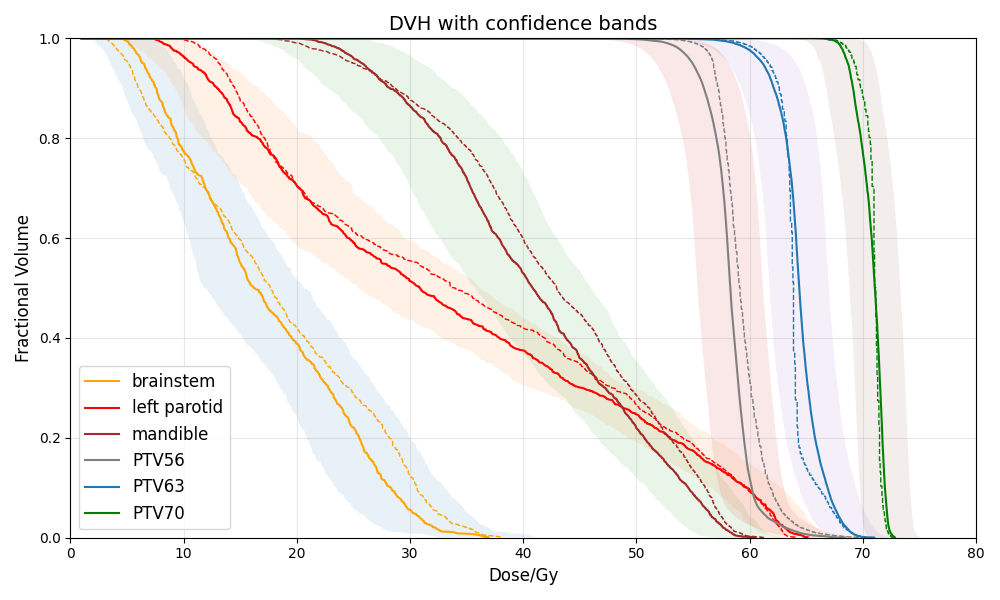}
  \caption{An example of the DVH of a patient in the Open-KBP dataset equipped with $95\%$ confidence intervals indicated as shaded regions. Dashed and solid curves pertain to the ground truth and model prediction respectively. The overall DVH score as defined in \cite{Babier} was 2.07 which was within the top 17 of the DVH score chart in \cite{Babier}.  }
  \label{fig:DVH_patient}
\end{figure}

Fig.~\ref{fig:DVH_patient} presents an illustrative DVH containing a number of organs-at-risk and target regions for an individual patient in the dataset of \cite{Babier}. Such a confidence band-enhanced DVH can be employed in the radiotherapy clinic to enable a more reliable interpretation of the model-predicted DVH.

\section{Discussion}
\label{sec:D}

We have refined and applied Deep Evidential Learning framework to radiotherapy dose prediction, our most prominent finding being that the uncertainty estimates obtained from model training inherited strong correlations with prediction errors. In the following, we conclude with a summary of key findings accompanied by a discussion of various limitations of our work.

\subsection{Summary of key findings}
\label{sec:summary}

The original loss function proposed in 
\cite{amini} did not work in this particular context of dose prediction. We found that regularizing the final layer with a sigmoid function and adding a mean-squared-error term led to a stable and effective implementation of Deep Evidential Learning. Upon completion of model training, we found that the model inherited uncertainty estimates at a more granular level compared to conventional MC Dropout and Deep Ensemble methods. Epistemic uncertainty ($U_{epis}$) in Deep Evidential Learning was highly correlated with prediction errors. Correlation indices (mutual information, Spearman's coefficients) were comparable or higher than those for MC Dropout and Deep Ensemble methods (see Table \ref{table:1}). 
Aleatoric uncertainty $(U_{alea})$ demonstrated a more significant shift in its empirical CDF upon addition of Gaussian noise to CT intensity distribution compared to $U_{epis}$ (see Fig.~\ref{fig:NoisePic}). It also displayed visibly weaker correlation with prediction errors relative to $U_{epis}$. These traits of $U_{alea}, U_{epis}$ appeared to be supportive of their conventional interpretations as reflecting data noise and model-related uncertainties respectively.

Another aspect of error-uncertainty relationship that appeared to distinguish Deep Evidential Learning 
was that the median error varied with uncertainty threshold much more linearly for $U_{epis}$ as compared to MC Dropout and Deep Ensemble methods, indicative of a more uniformly calibrated sensitivity to model errors (see Fig. \ref{fig:DE1}, \ref{fig:DE2}).
This is compatible with the fact that the uncertainty heatmaps generated by the model were found to be highly effective in identifying potential regions of model's inaccuracies.

\subsection{Towards incorporating our models in the radiotherapy clinic}

An essential plan evaluation tool for the radiation oncologist is the DVH that characterizes a dose distribution. 
We demonstrated how Eve's law of total variance enables one to express the predictive variance in terms
of $U_{alea}, U_{epis}$, and used this result to construct confidence intervals for DVH (see Fig.~\ref{fig:DVH_patient}). This technique can be harnessed by the radiotherapy treatment planning team to enable a more statistically informed interpretation of DVH in deep learning-guided treatment planning. In the Knowledge-Based-Planning pipeline, dose prediction models can be used as inputs to some dose mimicking model \cite{Babier2} that produces deliverable treatments via optimization of a set of objective functions \cite{Benson,Chan}.
The probabilistic 
dose distributions generated by our models can serve as robust inputs to the objective functions of \cite{Benson,Chan} to generate details of deliverable radiotherapy treatments. In this regard, it would be interesting to explore how the uncertainty estimates generated by our models translate to those of related quantities in the final treatment plan, such as the positions and motion dynamics of multileaf collimators and other machine parameters. This would pave the way for completing an uncertainty-aware AI-guided pipeline for radiotherapy treatment planning.

\subsection{Limitations and Future Directions}

Finally, we outline several limitations of our work together with corresponding suggestions for future directions. Although the simple 3D U-Net backbone architecture led to efficient and fast convergence, it would be interesting to study how more complicated network structures like the transformers-based models of \cite{L15,harry} perform after modifying their final layers so that there are four output heads corresponding to the parameters of $\mathcal{P}$ in Deep Evidential Learning. 

As a simple explicit example, a U-Net-like transformer model (Swin-UNet) was proposed in \cite{Swin2D} and \cite{Swin3D} for segmentation tasks. To enable Deep Evidential Learning on such backbone models, one can append a final 4-channel pointwise convolution layer to these transformer block-based architectures, which will yield the four parameters of the prior normal-inverse-gamma distribution in eqn. \eqref{fNIG} as model outputs. Model training should employ the maximum likelihood loss function of \cite{amini} or the refined version that we have proposed here in eqn. \eqref{loss}. 
More generally, this simple minimal prescription 
of appending a final layer of 4-channel pointwise convolution layer applies for other transformer-based vision models adapted for segmentation. If higher model complexity is needed, this final layer can be preceded by more multi-channel pointwise convolution layers, 
with other essential traits of transformer models left intact, e.g. multi-head self-attention modules, shifted window-based attention mechanism, etc. Details of these inner layers are unaffected and do not require additional modifications when encapsulating the transformer-based vision model within the framework of Deep Evidential Learning. 

Enabling Deep Evidential Learning on other backbone model structures will provide a more extensive study of the degree to which a low MAE can be attained while not compromising the uncertainty-error correlation. Indeed it will be interesting to explore how our refined Deep Evidential Learning model works for other regression tasks beyond the context of dose prediction, such as traffic forecasting in telecommunication networks in \cite{Moh1, Moh2}. It would be interesting to pursue whether the variety of model architectures examined in \cite{Moh1, Moh2} can be extended to uncertainty-aware ones using our techniques.

The MC Dropout and Deep Ensemble methods we studied for comparison purpose admit more complex variants which, in principle, may also yield aleatoric and epistemic uncertainties \cite{kendall,Toro,Egele}. It would be interesting to perform a similar analysis for them, although they would bring with them additional challenges (e.g. the use of Laplacian priors for MC Dropout setting as described in \cite{kendall}; computational cost in picking the optimal ensemble parameters as explained in \cite{Egele} where a novel automated approach for Deep Ensemble was proposed). 

We had used the same 
regularization term $
\mathcal{L}_{reg} = \left|
\log\left( \frac{y}{1-y}   \right) - \gamma
\right| (2\nu + \alpha)
$ in the loss function following
\cite{amini}, which is independent of the $\beta$ parameter of the prior distribution $\mathcal{P}$ in eqn. \eqref{fNIG}. It would be interesting to explore if there are other alternatives which express full dependence on all four parameters of $\mathcal{P}$. In an analogous framework for classification, Sensoy et al. in \cite{sensoy} showed that the regularization term measuring the Kullback-Leibler divergence from an uninformative prior (e.g. uniform distribution) worked well for classification problems, and notably, the authors of \cite{amini} had attempted to find the corresponding version for regression yet without success.  

Fundamentally, our modeling of dose prediction was based on the simplifying assumption that the patient's CT images and the oncologist's specification of the target regions are sufficient inputs for predicting the pareto-optimal dose distribution. In reality, effectiveness of radiation is often sensitive to a more complex web of biological factors. Nonetheless, we hope that our work can be a good starting point towards precision radiation oncology. Towards this goal, a pertinent future direction would be to expand the set of model's input features to include genetic, molecular and other unique clinical characteristics of each patient, beyond just using medical images.

\section{Conclusion}
\label{sec:C}

In this work, we have presented a novel application of Deep Evidential Learning in the domain of radiotherapy dose prediction. Using medical images of the OpenKBP Challenge dataset, we found that this model can be effectively harnessed to yield uncertainty estimates 
upon completion of network training. This was achieved only after reformulating the original loss function of \cite{amini} for a stable implementation. 
Since epistemic uncertainty was found to be highly correlated with prediction errors, its distribution could be used to discover and highlight areas of potential inaccuracies of the neural network, apart from being a diagnostic indicator of model reliability.  
Towards enhancing its clinical relevance, we demonstrated how to construct the predicted Dose-Volume-Histograms' confidence intervals. 
We hope that this work has furnished the crucial preliminary steps towards realizing Deep Evidential Learning for dose prediction models, paving another path towards quantifying the reliability of treatment plans in the context of knowledge-based-planning.

\section*{Declaration of competing interest}
All authors have no conflict of interest to declare.

\section*{Acknowledgments}
H.S.Tan wishes to acknowledge that the primary development of this work was conducted while he was affiliated with University of Pennsylvania, Perelman School of Medicine, Department of Radiation Oncology, to whom he is grateful for support.  





\bibliography{referenceStats}
\bibliographystyle{vancouver}

\end{document}